\documentclass[sigconf]{acmart}

\newcommand{\bX}{{\mathbf{X}}}
\newcommand{\bY}{{\mathbf{Y}}}
\newcommand{\bZ}{{\mathbf{Z}}}
\newcommand{\by}{{\mathbf{y}}}

\newcommand{\bx}{{\mathbf{x}}}
\newcommand{\bz}{{\mathbf{z}}}

\newcommand{\bs}{{\mathbf{s}}}

\AtBeginDocument{%
  }

\setcopyright{acmlicensed}
\copyrightyear{2024}
\acmDOI{}

\acmConference[3rd EAI Workshop SIGKDD'2024]{}{August 25 2024}{Barcelona, Spain}

\acmISBN{}

\acmSubmissionID{16}

\usepackage{multicol}
\usepackage{graphicx}
\usepackage{amsmath}
\usepackage{braket}
\usepackage{algpseudocode, algorithm}
\usepackage[noend]{algcompatible}
\DeclareMathOperator{\Tr}{Tr}

\begin{document}

\title{A Semidefinite Relaxation Approach for Fair Graph Clustering}

\author{Sina Baharlouei}
\authornote{Both authors contributed equally to this research.}
\email{sbaharlouei@ebay.com}
\affiliation{%
  \institution{eBay Search Ranking and Monetization}
  \city{San Jose, CA}
  \country{United States of America}
}

\author{Sadra Sabouri}
\authornotemark[1]
\email{sabourih@usc.edu}
\affiliation{%
  \institution{CS Department, University of Southern California}
  \city{Los Angeles, CA}
  \country{United States of America}
}



\begin{abstract}
Fair graph clustering is crucial for ensuring equitable representation and treatment of diverse communities in network analysis. Traditional methods often ignore disparities among social, economic, and demographic groups, perpetuating biased outcomes and reinforcing inequalities. This study introduces fair graph clustering within the framework of the disparate impact doctrine, treating it as a joint optimization problem integrating clustering quality and fairness constraints. Given the NP-hard nature of this problem, we employ a semidefinite relaxation approach to approximate the underlying optimization problem. For up to medium-sized graphs, we utilize a singular value decomposition-based algorithm, while for larger graphs, we propose a novel algorithm based on the alternative direction method of multipliers. Unlike existing methods, our formulation allows for tuning the trade-off between clustering quality and fairness. Experimental results on graphs generated from the standard stochastic block model demonstrate the superiority of our approach in achieving an optimal accuracy-fairness trade-off compared to state-of-the-art methods. 
\end{abstract}





\maketitle

\color{black}
\section{Introduction}

As machine learning algorithms increasingly influence critical decision-making tasks, from hiring~\citep{van2021machine} and lending~\citep{hall2021united} to law enforcement~\citep{washington2018argue} and healthcare~\citep{ghareh2023engineered, ashrafi2024effect}, it is crucial to diagnose and mitigate existing biases against protected groups. Unfair algorithms can lead to discrimination against marginalized groups, reinforcing systemic inequalities and undermining public trust in technology~\citep{dwork2012fairness}. While fairness in supervised learning has been extensively studied for more than a decade~\citep{zemel2013learning}, research on fairness in unsupervised learning tasks, particularly in graph clustering and community detection, is still limited~\citep{chierichetti2017fair}. Traditional clustering algorithms often optimize for overall performance without considering the distribution of benefits or harms across different demographic groups, leading to biased outcomes that unfairly disadvantage certain populations~\citep{bera2019fair}. For instance, in applications like customer segmentation~\citep{kansal2018customer}, biased clustering can result in unequal access to services or opportunities. Ensuring fairness in clustering can help mitigate these issues by promoting equitable treatment and inclusivity, which is particularly important in diverse and multi-faceted societies. Within the context of unsupervised learning tasks on graphs, the aim of fair graph clustering is to promote diversity within clusters (communities) while respecting existing friendships and connections~\citep{ziko2021variational}. For example, when dividing students in a class into different groups, traditional community detection focuses on creating groups where members have close connections. However, it is also desirable to ensure that these groups are diverse in terms of race and gender~\citep{GhodsiSN24}. Incorporating fairness into graph clustering can help create more balanced and representative communities, promoting inclusivity and equity in various applications~\citep{bera2019fair}.

In supervised learning, common notions of fairness can be divided into two main categories: group and individual fairness. Two primary concepts in the context of group fairness are statistical parity~\citep{dwork2012fairness} and equalized odds~\citep{hardt2016equality}. Statistical parity demands that the decision outcomes are independent of sensitive attributes such as race, gender, or age, aiming for an equal distribution of positive outcomes across these groups. Equalized odds, on the other hand, requires that the model's predictions, conditional on each class label, be independent of the sensitive attributes. Another important notion is individual fairness~\citep{gajane2017formalizing}, which is based on the idea that similar individuals should receive similar outcomes.
From a methodological perspective, the algorithmic fairness literature for supervised learning can be divided into three categories. Pre-processing methods~\citep{zemel2013learning} transform data into a fair representation under the given notion of fairness before the training phase, allowing the subsequent training to proceed on the transformed data as usual. Post-processing approaches~\citep{hardt2016equality, alghamdi2022beyond}, on the other hand, adjust the decision boundary of the final model after the training stage. A more resilient approach is to optimize accuracy and fairness constraints (or regularizers) jointly during the training procedure~\citep{zafar2017fairness}. In-processing methods~\citep{zafar2017fairness, baharlouei2023f} potentially achieve better accuracy-fairness tradeoffs. However, they are more challenging from an optimization standpoint, and standard optimization algorithms might struggle with convergence or scalability~\citep{lowy2022stochastic}.

Individual fairness notions for clustering are designed based on the idea that similar individuals must receive (almost) the same treatment~\citep{mahabadi2020individual}. In particular,~\citet{jung2020service} states within the context of the k-center problem that the distance of any data point from its assigned center must not be higher than $\alpha > 1$ times \textbf{its minimum radius}. The minimum radius is defined as the minimum radius of a ball containing at least $\lfloor \frac{n}{k} \rfloor$ of data points. Different methods within individual fair clustering use similar notions, but they might have different distance measures (e.g., k-means or k-median) or alternative definitions for the minimum radius~\citep{negahbani2021better}. A common issue with these methods is that they are primarily designed for center-based clustering problems (k-center, k-means, and k-median), and they are not applicable to graph clustering where the centers are not defined explicitly. Furthermore, promoting individual fairness can be in apparent contrast with the group fairness notions such as Balance~\citep{binns2020apparent}, that are more popular, more widespread, and have tangible applications~\citep{chierichetti2017fair,  backurs2019scalable}.

Applying group fairness notions to clustering problems is considered more challenging because ground-truth labels are not available during the training phase, making it impossible to evaluate fairness measures and criteria while learning to cluster~\citep{ziko2021variational}. Most group fairness criteria, such as equalized odds, equality of opportunity~\citep{hardt2016equality}, and sufficiency~\citep{baumann2022enforcing} in supervised learning, rely on access to labels. Consequently, measures for fair clustering are largely inspired by fairness notions from supervised learning that do not depend on labels~\citep{chierichetti2017fair}. Specifically, under the doctrine of demographic parity, \citet{chierichetti2017fair} proposes the balance measure, defined as the minimum ratio of two different protected groups across all clusters. For example, if a sensitive attribute labels each data point as either "man" or "woman," the clusters are considered balanced (fair under the demographic parity doctrine) if the ratio of women to men is the same in all clusters. ~\citet{backurs2019scalable} provides a near-linear time approximation to find fairlets that are small pre-clusters that fair. \citet{bera2019fair}, on the other hand, is a post-processing approach, making clusters fair through a reassignment policy.

The aforementioned methods and their variations are either pre-processing or post-processing approaches. They first learn fair representations (e.g., fairlets) and then apply the clustering algorithm to these fair representations, or they perform reassignments after the initial (unfair) clustering. These methods tend to result in loose approximations compared to scenarios where fairness and accuracy criteria are jointly optimized, similar to in-processing methods in supervised learning.

Alternatively, \citet{baharlouei2020renyi} handles fairness constraints during the learning phase. To ensure a convergent algorithm, \citet{baharlouei2020renyi} updates the fairness parameters after every update of data point assignments to the $k$ clusters, making it only applicable to very small datasets. 
Moreover, ~\citet{GhodsiSN24} proposes a non-negative matrix factorization method to promote individual fairness in graph clustering. Similar to~\citet{baharlouei2020renyi}, their approach does not provide a convergence guarantee for the proposed algorithm. Moreover, as we observed in practice, the optimization procedure is unstable and can lead to solutions that are either completely unfair (equivalent to no fairness constraints) or all nodes will be in a single cluster (no clustering).  

Our approach, however, handles fairness criteria by adding it as a regularization term to the objective function. This allows for a desired tradeoff between fairness and clustering quality by adjusting the parameter for the fairness term. Furthermore, adding the fairness regularization term in our framework, does not affect the convergence of the proposed algorithm.

\noindent \textbf{Contributions:} We devise a framework based on semi-definite programming relaxation to formulate the fair graph clustering under the demographic parity doctrine in a tractable optimization problem (Section~\ref{sec: Method}). Moreover, we propose two algorithms (Section~\ref{sec: Algorithms}) to solve the proposed optimization problem (eigenvalue truncation and alternative direction method of multipliers) in a scalable fashion. Our methodology allows the user to choose the level of desired fairness by tuning a hyper-parameter balancing the importance of fairness and clustering quality. Therefore, unlike the majority of methods in the literature, one can have access to multiple clustering schema in the fairness-accuracy tradeoff curve. Furthermore, to evaluate our approach, we propose a measure based on the area under the fairness-accuracy tradeoff curve. This measure can be used in both scenarios where ground-truth cluster assignments are available and unknown. Our experiments demonstrate the superiority of our approach in terms of fairness-accuracy tradeoff compared to the state-of-the-art approaches supporting graph clustering.
\begin{figure*}[ht]
    \centering
    \vspace{-5mm}
    \begin{multicols}{3}
        \includegraphics[width=0.3\textwidth]{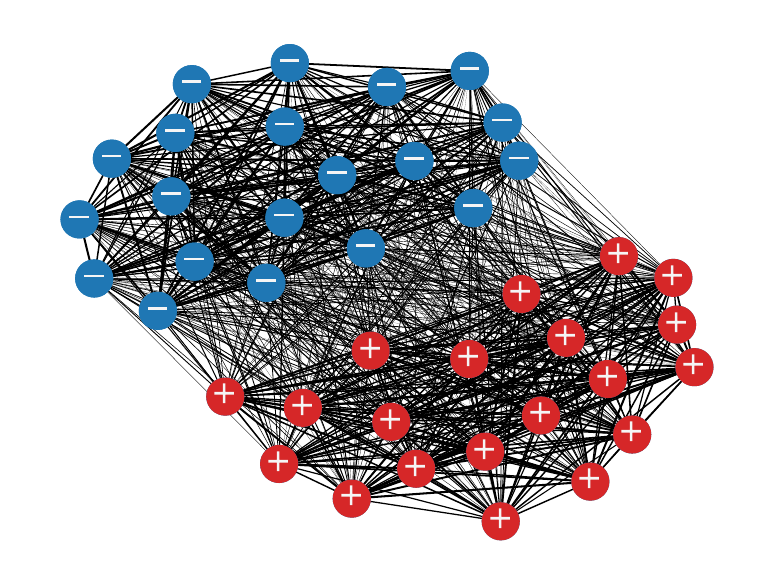}\par 
        \includegraphics[width=0.3\textwidth]{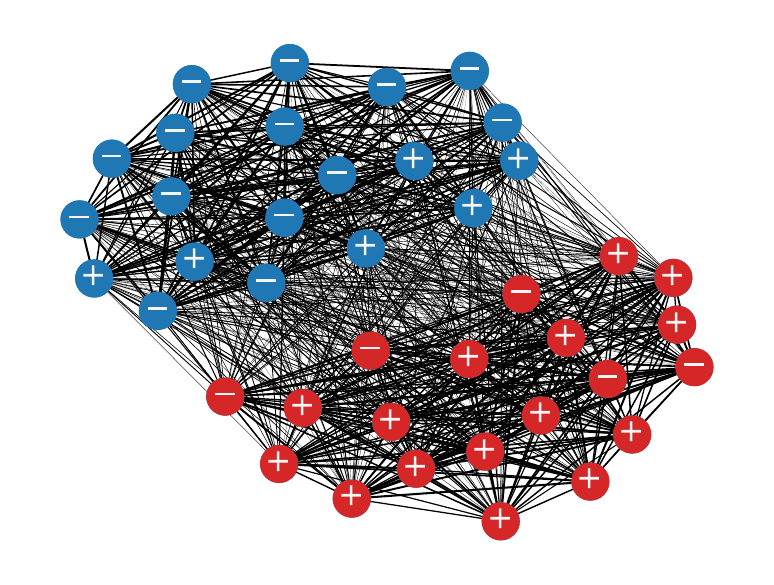}\par
        \includegraphics[width=0.3\textwidth]{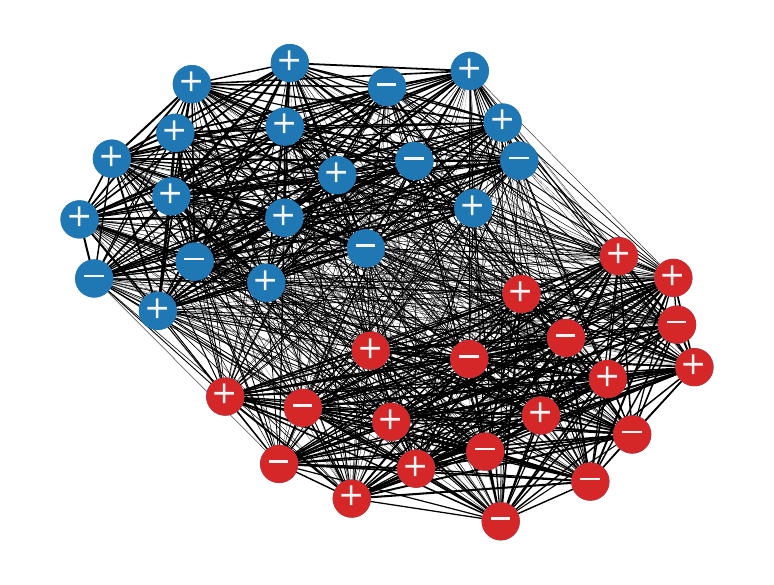}
    \end{multicols}
    \begin{multicols}{3}
        a)\includegraphics[width=0.3\textwidth]{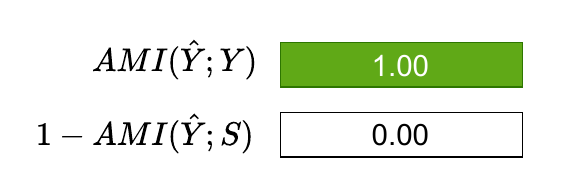}\par 
        b)\includegraphics[width=0.3\textwidth]{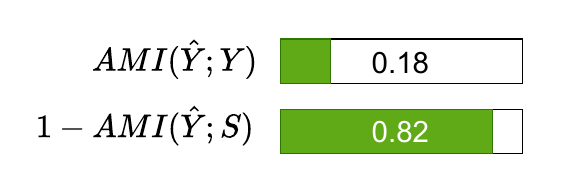}\par
        c)\includegraphics[width=0.3\textwidth]{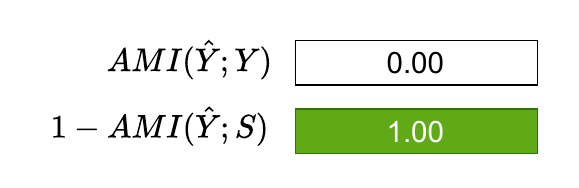}
    \end{multicols}
    \caption{An example of weighted imbalanced graph clustering. Node's specificity feature is indicated with colors, i.e., \textcolor{red}{red} and \textcolor{blue}{blue}, an edge boldness is proportional to its weight. Plus, the edge's length is inversely proportional to its weight. Node labels, i.e $\{-,+\}$, shows the predicted cluster. $AMI$ is Adjusted Mutual Information~\cite{vinh2009information} used to evaluate the predicted labels, $\hat{Y}$, to real temporal clusters, $Y$, and specificity $S$. a) The predicted labels for clusters are all based on temporal information. The clustering method ignores fairness and predicts clusters only based on cluster locations. c) Clustering is solely based on fairness criteria. b) A trade-off between the clustering and fairness (temporal vs. specificity features)}
    \label{fig:casestudy-tradeoff}
\end{figure*}

\section{Formulation and Methodology}
\label{sec: Method}
In this section, we first introduce the notation and assumptions on the clustering problem. Next, we state the fair graph clustering formulation and its SDP relaxation.
\subsection{Notation}
We refer to $X = \{\bx_1, \cdots, \bx_n\}$ as the set of $n$ data objects to be clustered into several (k) partitions. 
Each data object $\bx_i$ is labeled by an unknown latent variable $y_i \in \{1, \cdot, k\}$ representing the actual (ground-truth) cluster (or community) assigned to data $i$. Given $k$ as the number of underlying communities, and $n$ data points, one can form a matrix $\bY^{n\times k}$ with $\{ij\}$-th entry equals $1$ if data point $i$ belongs to the community $j$ and $0$ otherwise. 
Similarly, each data object is labeled by $s_i$ as its specificity indicator, determining to what protected group data point $i$ belongs to (e.g., man or woman). 
$w: X \times X \to [0, 1]$ is a weight function which shows similarity between objects. We assume that as $w(x_i, x_j)$ is higher, $x_i$ is more similar to $x_j$. Also $w(x_i,x_j)= w(x_j,x_i)$ for all $i$ and $j$s.
We construct the clustering graph as a weighted graph $G=(V, E)$ with $V = \{v_i: \forall i \in [n]\}$, $E = \{ \{v_i, v_j\} : w(x_i, x_j) > 0 \}$ and a weight function $W$ such that $W(v_i, v_j) = w(x_i, x_j)$. We define matrix $A$ as a symmetric matrix representing weights on edges, i.e. $A_{ij} = W(v_i, v_j)$. Note that, in problems where $\bX$ is given (no adjacency matrix), the adjacency matrix can be established by either thresholding on the distance of data points or assigning $W(v_i, v_j)$ as a decreasing function of data point distances (one choice can be $W(v_i, v_j) = \frac{1}{d(\bx_i, \bx_j)}$, where $d(., .)$ represents the Euclidean distance between two data points. 

We use the standard stochastic block model setting for characterizing the graph structure associated with the community detection problem. Assume that Let $\Psi^{k\times k}$ be the probability matrix whose $ij$-th entry represents the probability of a member of $i$-th community to be connected to a member of $j$-th community. A widely used special case is when $\Psi_{ii} = p \gg \Psi_{ij} = q$, which means if two points are within the same community, they have a much higher chance (p) to be connected compared to the case when they are from different communities (q). Assume that $A^{n\times n}$ is the adjacency matrix of graph $\mathcal{G}$. 
\subsection{SDP Relaxation for Fair Clustering Problem}
We start by formulating the maximum likelihood problem for community detection under the stochastic block model. Define:
\begin{equation*}
    M = Y \Psi Y^T
\end{equation*}
Note that we do not have access to $Y$ and $\Psi$, and both of these matrices are latent variables that can be learned during the clustering procedure. The log-likelihood function, in this case, is defined as:
\begin{equation}
    \ell(Y, \Psi) = \sum_{i < j} A_{ij} \log(M_{ij}) + (1 - A_{ij})\log(1 - M_{ij})
\end{equation}
As shown in~\citet{amini2018semidefinite}, the maximum likelihood estimator can be stated as:
\begin{equation}
    \max_{Y} Y^T A Y - \lambda Y^T \mathbf{1} \mathbf{1}^T Y 
\end{equation}
where the second term represents a balanced constraint, making sure the size of the clusters is close to each other. 

For the simplicity of the presentation, we consider the simple case where the matrix $\bY$ and sensitive attributes ($\bs$) are binary vectors ($2$ communities and $2$ levels for the sensitive attribute). We discuss how to generalize the algorithms to $k$ communities and multiple (non-binary) sensitive attributes in Appendix~\ref{appendix: multiple}. Therefore, the maximum likelihood problem for finding two communities can be written as:
\begin{equation}    
    \begin{aligned}
    \max_{\by} \quad & \by^T A \by\\
    \textrm{s.t.} \quad & \by \in \{-1,1\}^n \quad \by^T \mathbf{1} = 0
    \end{aligned}
    \label{eq:clustering-problem-1}
\end{equation}
As we mentioned earlier, the clustering model is fair if and only the number of each protected group level in all clusters is the same. As a result, the clustering is fair under the demographic parity doctrine, if and only $\by^T \bs = 0$~\citep{kleindessner2019guarantees}. Thus, the Fair Graph Clustering optimization can be formulated as:
\begin{equation}    
    \begin{aligned}
    \max_{\by} \quad & \by^T A \by\\
    \textrm{s.t.} \quad & \by \in \{-1,1\}^n \quad \by^T \mathbf{1} = 0 \quad \by^T \bs = 0,
    \end{aligned}
    \label{eq:fair_clustering}
\end{equation}
This problem is NP-hard due to the binary constraint on the entries of $\by$. Hence, it is not tractable in the current format. We apply the SDP relaxation technique~\citep{amini2018semidefinite} to reformulate the above problem. 

Note that $\by^T A \by$ is a scalar. Therefore, $\by^T A \by = \Tr(\by^TA\by) = \Tr (A\by\by^T)$. Now, let $\bZ = \by \by^T$. As a result, Problem~\eqref{eq:fair_clustering} is equivalent to:
\begin{equation}    
    \begin{aligned}
    \max_{Z} \quad & \Tr(AZ)\\
    \textrm{s.t.} \quad & \by \in \{-1,1\}^n \quad Z = \by\by^T \quad \by^T 1 = 0 \quad \by^T \bs = 0
    \end{aligned}
    \label{eq:clustering-problem-eq}
\end{equation}
Since $Z = \by \by^T$, it implies that $Z$ is a rank-1 symmetric matrix. Moreover, since $\by \in \{-1, 1\}^n$, we have $Z_{ii} = 1$. Therefore, Problem~\eqref{eq:fair_clustering} can be equivalently formulated as:
\begin{equation}    
    \begin{aligned}
    \max_{Z} \quad & \Tr(AZ)\\
    \textrm{s.t.} \quad & \quad Z_{ii} = 1 \quad \textrm{Rank}(Z) \leq 1 \quad \by^T 1 = 0 \quad \by^T \bs = 0 \quad  Z \succeq 0
    \end{aligned}
    \label{eq:fair_clustering-problem-eq}
\end{equation}
Since the Ranking constraint is non-convex and intractable, we \textbf{relax} the constraint to a nuclear norm constraint (SDP relaxation) as follows:
\begin{equation}    
    \begin{aligned}
    \max_{Z} \quad & \Tr(AZ)\\
    \textrm{s.t.} \quad & \quad Z_{ii} = 1 \quad \|Z\|_{*} \leq 1 \quad \by^T 1 = 0 \quad \by^T \bs = 0 \quad  Z \succeq 0
    \end{aligned}
    \label{eq:fair_clustering-SDP}
\end{equation}
Since the objective function and all constraints are convex, and the problem is convex \textbf{maximization}, it always has an answer on the \textbf{boundary}. Therefore, we can convert $||Z||_* \leq 1$ to $||Z||_* = 1$. Furthermore, $\by^T \mathbf{1} = 0$ is equivalent to $\|\by^T \mathbf{1}\|_2^2 = 0$ which means $\mathbf{1}^T Z \mathbf{1}^T = 0$. Similarly, $\bs^T Z \bs = 0$ for the fairness constraint. Therefore, the SDP-relaxed problem can be written solely based on the optimization parameter $Z$ as follows:
\begin{equation*}    
    \begin{aligned}
    \max_{Z} \quad & \Tr(AZ)\\
    \textrm{s.t.} \quad & \quad Z_{ii} = 1 \quad \|Z\|_{*} \leq 1 \quad \mathbf{1}^T Z \mathbf{1} = 0 \quad \bs^T Z \bs = 0 = 0 \quad  Z \succeq 0
    \end{aligned}
    \label{eq:fair_clustering-SDP_Z}
\end{equation*}
We bring two zero-equality constraints to the objective function as regularization terms with $\lambda$ and $\mu$ coefficients, respectively. Also, matrix $Z$ is a positive semi-definite matrix (PSD). Therefore, the problem is formulated as:
\begin{equation}    
    \begin{aligned}
    \max_{Z} \quad & \Tr\Big((A - \mu \mathbf{1}\mathbf{1}^T - \lambda\bs \bs^T)Z\Big)\\
    \textrm{s.t.} \quad & \quad Z_{ii} = 1 \quad \|Z\|_{*} \leq 1 \quad  Z \succeq 0.
    \end{aligned}
    \label{eq:fair_clustering-soft}
    \tag{FAIR-SDP}
\end{equation}


\section{Algorithms for Solving Fair SDP-Relaxed Fair Clustering}
\label{sec: Algorithms}
We propose two algorithms for solving~\ref{eq:fair_clustering-soft}. The first one is a variation of spectral clustering applied on $\Tilde{A} = A - \mu \mathbf{1}\mathbf{1}^T - \lambda \bs \bs^T$. In the case of binary clustering, we find the eigenvector corresponding to the second largest eigenvalue~\citep{sarkar2011community}, and we assign clusters based on the sign of the eigenvector entries. The procedure is presented in Algorithm~\ref{alg: svd}. We discuss how we can generalize this algorithm to the case of multiple sensitive attributes and multiple (k > 2) clusters in Appendix~\ref{appendix: multiple}. 
\begin{algorithm}
    \caption{Fair Graph Clustering via SVD}
    \label{alg: svd}
    \begin{algorithmic}[1]
	 \STATE \textbf{Input}: fairness importance weight $\lambda$,  cluster size balance parameter $\mu$, Adjacency Matrix $A$, sensitive attribute vector $\bs$. 
    \STATE Set $\Tilde{A} = A - \mu \mathbf{1}\mathbf{1}^T - \lambda \bs \bs^T$
    \STATE Let $\Tilde{A} = U \Sigma V^T$ \: \: \: \: (Singular Value Decomposition)
    \STATE Let $\mathbf{v}_2 = \textrm{col}_2 (V)$
    \STATE $\textrm{cluster} (i) = \textrm{sign}(\mathbf{v}_2[i]) \quad i = 1, \dots, n$
\end{algorithmic}
\end{algorithm}
The algorithms based on singular value decomposition are criticized in the literature~\citep{GhodsiSN24} due to their scalability issues. Algorithm~\ref{alg: svd} needs $\mathcal{O}(n^3)$ operations where $n$ is the number of data points to be clustered. Therefore, it might not be feasible to apply such an algorithm to large-scale graphs. Alternatively, we offer an alternative direction method of multipliers (ADMM) to perform fair graph clustering on~\eqref{eq:fair_clustering-soft}. The details are presented in Algorithm~\ref{alg: admm} (see Appendix~\ref{appendix: ADMM}). 
\section{Numerical Results}
In this section, we designed several numerical experiments based on the stochastic block model to present our method's preliminary results for fair graph clustering. 

In the first experiment, for the sake of illustration, we generate a random weighted graph with two clusters with $20$ and $10$ nodes within each cluster. We connect two points within each cluster with a probability randomly sampled from $[0.5, 1.0]$, and for each of two nodes in different clusters, we connect with a probability randomly sampled from $[0, 0.5]$. As a result, we have significantly stronger connections within clusters than between clusters. Figure~\ref{fig:casestudy-tradeoff} shows an example of this randomly generated graph and is plotted using networkx~\citep{hagberg2008exploring}. 
We generate the sensitive attribute vector (binary with two colors, red and blue) for graph nodes by randomly sampling a random variable $p \sim Bernoulli(0.5)$. With this, we would have an equally distributed specificity for our nodes in the graph. Specified node values are shown in Figure~\ref{fig:casestudy-tradeoff} with red and blue colors.

Since we do have access to the ground-truth cluster assignments, we can utilize clustering quality metrics such as Adjusted Mutual Information (Figure~\ref{fig:casestudy-tradeoff}) and V-measure implemented in scikit-learn~\citep{pedregosa2011scikit}. Adjusted Mutual Information~\cite{vinh2009information} ($AMI(Z;\hat{Z})$) is the amount of information revealed about the random variable $\hat{Z}$ (predicted clusters) when knowing the value of the random variable $Z$ (true clusters). The higher the adjusted mutual information, the higher the clustering quality (closer to the ground truth). When increasing $\lambda$, i.e., increasing the algorithm's fairness, adjusted mutual information between the sensitive attribute vector and the predicted clusters increases. That means predicted clusters are getting better representations of specificity (sensitive attributes) than the actual clusters. We call $AMI(\hat{Y};Y)$ \textbf{temporal AMI} (clustering quality compared to the ground truth) and $AMI(\hat{Y};S)$ \textbf{specificity AMI} (how fair the predicted clustering assignments are).

We applied our method (both Algorithm~\ref{alg: svd} and Algorithm~\ref{alg: admm} return the same results) for fair clustering to this graph. We have two tuning hyper-parameters. $\mu$ controls the balance between classes and prevents the algorithm from clustering all nodes as one cluster. $\lambda$ controls the algorithm's fairness, as $|\lambda|$ is greater the algorithm put higher importance to fairness. From \ref{fig:casestudy-tradeoff}-a to c, we changed the $\lambda$ from $0$ to $-0.4$ while keeping $\mu=1$.


\begin{figure*}[ht]
    \centering
    \begin{multicols}{3}
        a)\includegraphics[width=0.3\textwidth]{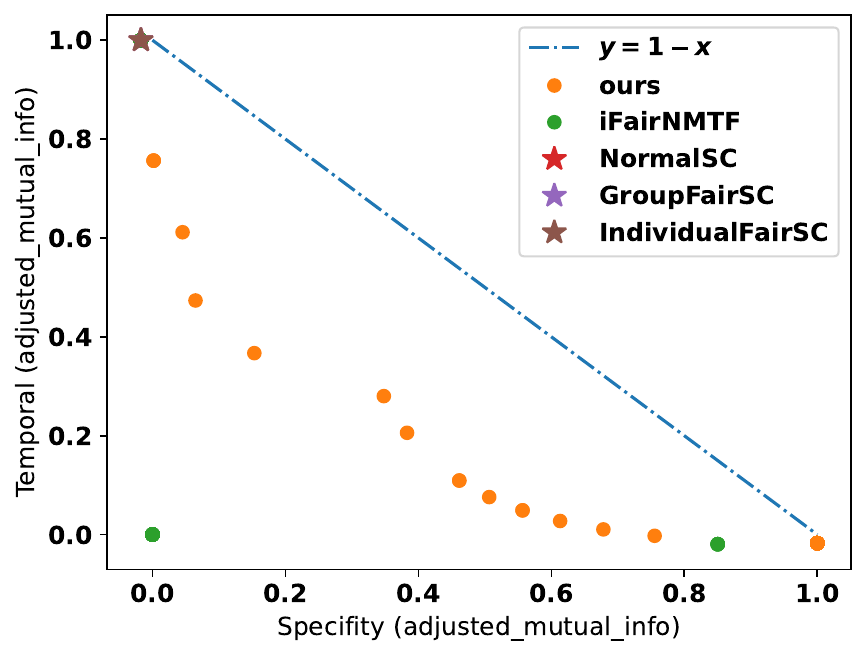}\par 
        b)\includegraphics[width=0.3\textwidth]{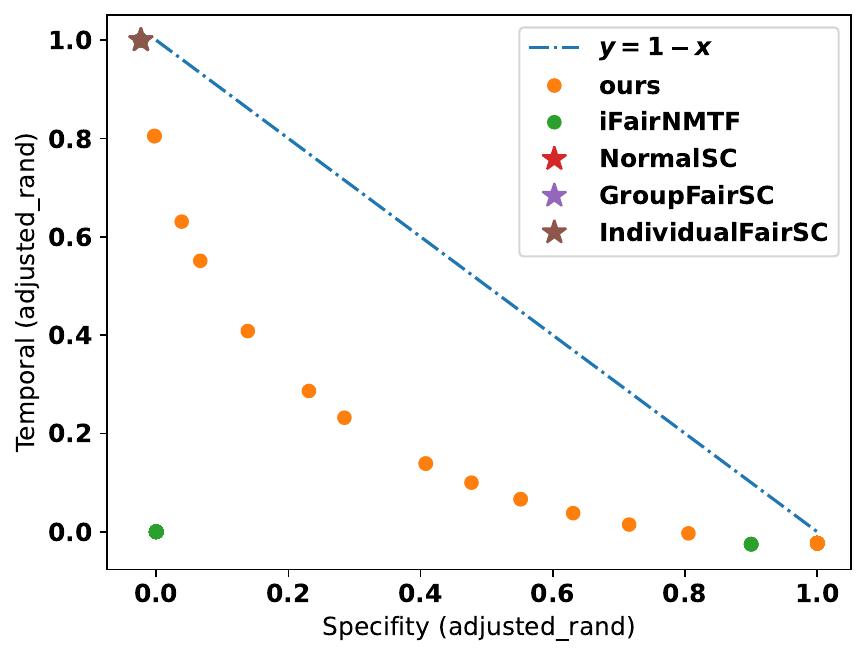}\par
        c)\includegraphics[width=0.3\textwidth]{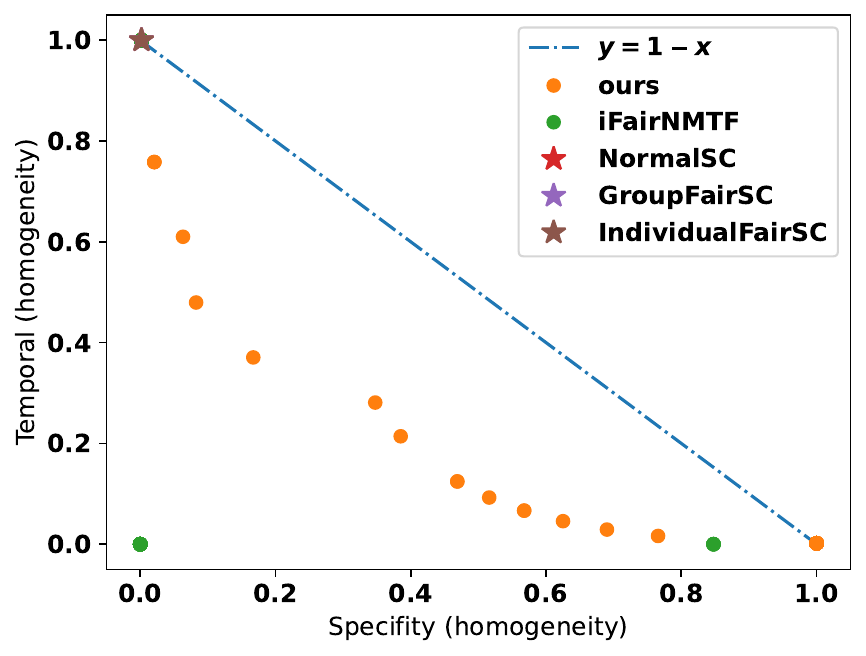}
    \end{multicols}
    \caption{Clustering Accuracy and fairness trade-off for our model, iFairNMT~\cite{GhodsiSN24}, iFarSC~\cite{gupta2022consistency} (IndividualFairSC), GroupFairSC~\cite{GhodsiSN24} and NormalSC~\cite{GhodsiSN24} in a) adjusted mutual information~\cite{vinh2009information}, b) adjusted rand index~\cite{chacon2023minimum} and c) v measure~\cite{rosenberg2007v} using $1000$ samples.}
    \label{fig:experiment-compare}
\end{figure*}
Next, we designed an experiment to compare our method with existing state-of-the-art approaches for fair graph clustering. We generated a balanced random graph with two clusters, with $1000$ nodes in each cluster. Building adjacency matrix $A$, we sample a random variable $p \sim Bernoulli(0.05)$ for between cluster node edges and $q \sim Bernoulli(0.90)$ for within-cluster node edges (standard stochastic block model setting). With this, we ensured the existence of clusters while keeping the graph generation process randomized. We compare our method with several methods supporting graph clustering: iFairNMT~\cite{GhodsiSN24}, iFarSC~\cite{gupta2022consistency} (IndividualFairSC), GroupFairSC~\cite{GhodsiSN24} and NormalSC~\cite{GhodsiSN24}. Alongside their fair clustering algorithm, ~\citet{GhodsiSN24} implemented three other algorithms for clustering (iFairNMT, GroupFairSC, NormalSC). We used their implementations as our baseline. We run these algorithms with different hyper-parameters to obtain the accuracy-fairness tradeoff curve. For all methods, we computed adjusted mutual information, adjusted rand index, and $v$ measure for temporal (goodness of clustering) and specificity (conforming fairness criteria). The results are depicted in Figure~\ref{fig:experiment-compare}. 
IfairNMT~\citep{GhodsiSN24} mostly clustered all the nodes into one cluster, which is not a suitable clustering, and set the AMI for both temporal and specificity to $0$. In other cases, the clusters provided by ifairNMT were actually worse than ours both in terms of clustering quality and fairness. 


Furthermore, in Appendix~\ref{appendix: hyper}, we discuss how to choose hyper-parameters for balancing the size ($\mu$) and fairness coefficient ($\lambda$) to obtain the best possible accuracy-fairness tradeoff. Also, several points regarding the implementation of the algorithms and further experiments on the performance of our method in different settings can be found in the same appendix. All implementations are available at \url{https://github.com/sadrasabouri/sdp-fair-graph-clustering}.

\textbf{Future Directions:} As future directions, we will expand our theory and applications to continuous sensitive attributes. Moreover, we explore the possibility of extending our algorithms to directed graphs as they pose more complex constraints to the underlying optimization problem. Further, we extend our experiments to large-scale real datasets consisting of multiple sensitive attributes.


\bibliographystyle{ACM-Reference-Format}
\bibliography{KDD}

\clearpage

\appendix
\section{Extension of Fair Graph Clustering Formulation to Multiple 
Sensitive Attributes and Multiple Clusters (Communities)}
\label{appendix: multiple}
In this section, we first discuss how to extend Algorithm~\ref{alg: svd} to $k > 2$ clusters. Next, we talk about extending the formulation to multiple sensitive attributes.

\noindent  \textbf{Clustering For More Than 2 Clusters.} There are two common methods for extending Algorithm~\ref{alg: svd} to multiple clusters. 
\begin{enumerate}
    \item We start with two clusters, then based on a clustering metric like homogeneity, decide to sub-cluster which of those two clusters. We will go on until we reach $k$ clusters.

    \item instead of finding the second largest singular vector of matrix $\Tilde{A}$, we define $\Tilde{L} = \textrm{diag}(\Tilde{A}) - \Tilde{A}$. Then, we find the first $k$ eigenvectors corresponding to the $k$ smallest eigenvalues of $\Tilde{L}$ and truncate others. Then, we apply a simple $k$-means algorithm (Lloyd's Algorithm, for instance) to the obtained truncated matrix to find the final $k$ clusters. 
\end{enumerate}

\noindent \textbf{Multiple Sensitive Attributes.} A natural extension of \eqref{eq:fair_clustering-soft} to multiple sensitive attributes is to consider a separate vector for each level of sensitive attributes. For instance, assume that we have a race attribute with three different values and gender with two values. Therefore, we have $6$ possible combinations. In this case, we define $\bs_1$ to $\bs_6$ whose entries determine whether a data point belongs to each one of these 6 combinations (therefore, for data point $i$, only one of $i$-th entries among $\bs_1$ to $\bs_6$ is $1$ and the rest are zero). Thus, instead of $\lambda \|\bs^T \by\|_2^2$, we can add $\sum_{i=1}^6 lambda_i \|\bs^T \by\|_2^2$ as the fairness regularization term to the objective function. If all protected groups have the same priority, we set all $\lambda_1$ to $\lambda_6$ to the same value. Otherwise, we put more weight on the protected groups with higher priorities.

\section{Deriving ADMM Update Rules for Fair Graph Clustering}
\label{appendix: ADMM}
We start by defining $B = - \Tilde{A} = - A + \mu \mathbf{1}\mathbf{1}^T + \lambda\bs \bs^T$. Therefore \eqref{eq:fair_clustering-soft} can be rewritten as:
\begin{equation}    
    \begin{aligned}
    \min_{Z} \quad & \Tr\Big(BZ\Big)\\
    \textrm{s.t.} \quad & \quad Z_{ii} = 1 \quad \|Z\|_{*} \leq 1 \quad  Z \succeq 0.
    \end{aligned}
    \label{eq:fair_clustering-B}
\end{equation}

Let $\bz$ be the main diagonal of matrix $Z$. Note that the nuclear norm is used as a surrogate for the non-convex non-smooth rank function. We add this term to the objective function via a Lagrangian multiplier $\beta$. Moreover, we define an auxiliary variable $P = Z$, as handling all constraints on $Z$ simultaneously is not possible. Therefore, the optimization problem turns to:
\begin{equation}    
    \begin{aligned}
    \min_{Z, P} \quad & \Tr\Big(BZ\Big) + \beta \|P\|_{*} + \langle \boldsymbol{\alpha}, \bz - \mathbf{1} \rangle + \frac{\rho}{2} \|\bz - \mathbf{1}\|^2 \\
    & + \langle \boldsymbol{\Gamma}, P - Z \rangle + \frac{\rho}{2} \|P - Z\|_F^2\\
    \textrm{s.t.} \quad & P \succeq 0.
    \end{aligned}
    \label{eq:fair_clustering-B2}
\end{equation}
Now, we consider two blocks ($P$ and $Z$), and we alternatively update them fixing the other one. The procedure is described in Algorithm~\ref{alg: admm}.
\begin{algorithm}
    \caption{Fair Graph Clustering via ADMM}
    \label{alg: admm}
    \begin{algorithmic}[1]
    
    \STATE \textbf{Input}: fairness importance weight $\lambda$,  cluster size balance parameter $\mu$, Adjacency Matrix $A$, sensitive attribute vector $\bs$, T: Number of iterations.

    \vspace{1mm}
    \STATE \textbf{Initialize }: $Z_0 = 0, P_0 = 0, \boldsymbol{\Gamma}_0 = \mathbf{1} \mathbf{1}^T, \boldsymbol{\alpha}_0 = \mathbf{1}$, .
    
    \FOR {$t = 1, \ldots, T$}
    \vspace{1mm}
    \STATE Set $Z_{i} = \textrm{argmin}_{Z} \Tr\Big(BZ\Big) + \langle \boldsymbol{\alpha}_{i-1}, \bz - \mathbf{1} \rangle + \frac{\rho}{2} \|\bz - \mathbf{1}\|^2 + \langle \boldsymbol{\Gamma}_{i-1}, P_{i-1} - Z \rangle + \frac{\rho}{2}  \|P_{i-1} - Z\|_F^2$

    \vspace{1mm}
    \STATE Set $P_{i} = \textrm{argmin}_{P} \: \: \beta \|P\|_{*} + \langle \boldsymbol{\Gamma}_{i-1}, P - Z_{i} \rangle + \frac{\rho}{2}  \|P - Z_{i-1}\|_F^2$


    \vspace{1mm}
    \STATE Set $\boldsymbol{\alpha}_{i} = \boldsymbol{\alpha}_{i-1} + \rho (\bz_{i-1} - \boldsymbol{1})$

    \vspace{1mm}
    \STATE Set $\boldsymbol{\Gamma}_{i} = \boldsymbol{\Gamma}_{i-1} + \rho (P_i - Z_i)$
    \ENDFOR

    \STATE Return $Z_T$, $P_T$ 
\end{algorithmic}
\end{algorithm}

\noindent \textbf{Remark 1:} Algorithm~\ref{alg: admm} is iterative based and does not need eigenvalue decomposition. Therefore, it is more suitable for large-scale. The main reason is Algorithm~\ref{alg: svd} requires singular value decomposition on the adjacency matrix, that is infeasible when $n$ is large ($\mathcal{O}(n^3)$ operations). Therefore, it is crucial to use Algorithm~\ref{alg: admm} in those cases, in which, per-iteration complexity is $\mathcal{O}(n^2)$.  

\noindent \textbf{Remark 2:} While Algorithm~\ref{alg: svd} finds the exact solution to the fair clustering problem, ADMM algorithm needs $\mathcal{O}(\frac{1}{\epsilon})$  to find an $\epsilon$-optimal solution to the problem. For small-size datasets, our experiments show the same results when the number of iterations is above $500$. As a future work, we do more experiments on larger datasets to illuminate the difference in performance and actual runtime of two algorithms in practice.    

\section{Hyper-parameter Tuning and Its Effect on Fairness-Accuracy Tradeoff}
\label{appendix: hyper}
\begin{figure}[ht]
    \centering
    \includegraphics[width=\linewidth]{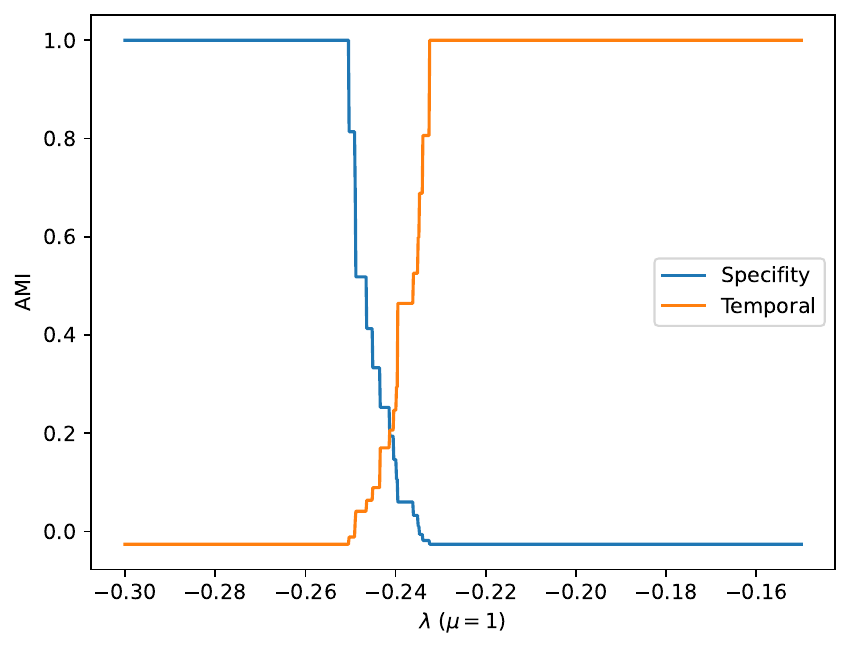}
    \vspace{-5mm}
    \caption{Specifity AMI and temporal AMI change per $\lambda$ when $\mu=1$ for 1000 samples. In this example, for $\lambda \leq -0.26$, the algorithm ignores the fairness; for $\lambda \geq -0.22$, the algorithm ignores the temporal information between nodes.}
    \label{fig:change_lambda_01_03_observe_AMI_spec_temp}
\end{figure}

\begin{figure}[ht]
    \centering
    \includegraphics[width=\linewidth]{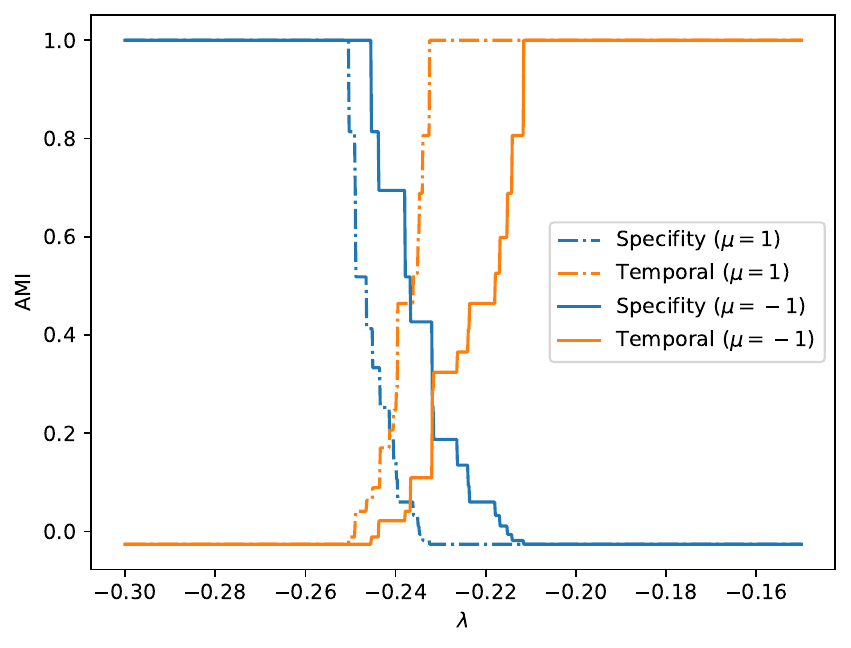}
    \caption{Change of specificity and temporal AMI per changing $\lambda$ when $\mu = \pm 1$ for 1000 samples. When $\mu = -1$, the trade-off range is wider than the case when $\mu = 1$.}
    \label{fig:change_lambda__03__015_mu_1VSmu1}
\end{figure}

In this section, we discuss the effect of choosing hyper-parameters $\mu$ (balance for size of clusters) and $\lambda$ (fairness importance coefficient).
\begin{figure}[ht]
    \centering
    \includegraphics[width=\linewidth]{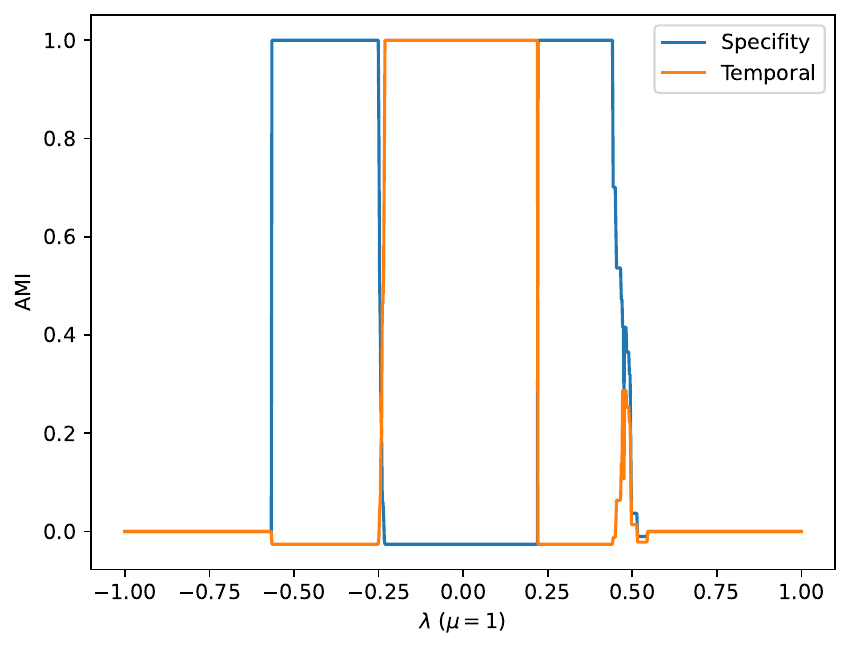}
    \caption{Specificity and Temporal AMI change for $\mu=1$ as a function of $\lambda$ ($1000$ samples). While the changes concerning the $\lambda$ are not symmetric, the change is smoother when $\lambda < 0$ compared to the sharper change for $\lambda > 0$. When $\lambda > 0.5$ or $\lambda < -0.6$, our method clusters all nodes as one cluster, resulting in a dump in both specificity and temporal AMIs.}
    \label{fig:change_lambda_-1_1_observe_AMI_spec_temp}
\end{figure}
\noindent \textbf{Effect of $\mu$ and $\lambda$ Signs} Since the constraints in~\eqref{eq:fair_clustering-problem-eq} are in the form of equality, the coefficients for adding them as soft constraints (regularization terms) to~\eqref{eq:fair_clustering-soft} is signed free (think of them as Lagrangian multipliers). During the experiments, we observed that our method's performance is dropped when $\lambda > 0$. In Figure~\ref{fig:change_lambda_-1_1_observe_AMI_spec_temp}, you can see the change in specificity and temporal AMI when $\lambda$ changes from -1 to 1 while $\mu=1$ is fixed. We have a binary trade-off for $0 < \lambda <0.3$, i.e., the clustering changes from solely temporal-based to solely specify-based. On the other hand, when $-0.5 < \lambda < 0$ we have a smooth trade-off near $\lambda = -0.25$. We had a zoomed-in version of this trade-off in Figure \ref{fig:change_lambda_01_03_observe_AMI_spec_temp}. This smooth trade-off zone makes sure the user can choose the amount of desirable fairness depending on the application. 
For the role of $\mu$ in the trade-off, we again found that $\mu = -1$ offers a wider range for specificity (fairness) and temporal (accuracy) trade-offs. In particular, Figure~\ref{fig:change_lambda__03__015_mu_1VSmu1} shows that the range for the trade-off is wider when $\mu = -1$ for $\lambda < 0$. This means that finding the trade-off range is easier, and there are more trade-off points when $\mu = -1$.
As a result, our initial findings suggest tuning $\mu$ and $\lambda$ in the negative zone, as it offers a smoother tradeoff curve with higher performance.

\noindent \textbf{Model Behavior for Large $\lambda$} From \eqref{eq:fair_clustering-soft}, we know $\lambda s s^T Z = \lambda (z^T s)^2$ accounts for the fairness term. When $\lambda > 0$ is large, $(z^T s)^2$ is the dominating term. Therefore, $z=-s$ is a clustering solely based on fairness. When $\lambda < 0 $ is large, , we try to maximize the $(z^T s)^2$ term, the model behaves similarly. Therefore, fairness accuracy trade-offs always happens near $0$.
\begin{figure}[ht]
    \centering
    \includegraphics[width=\linewidth]{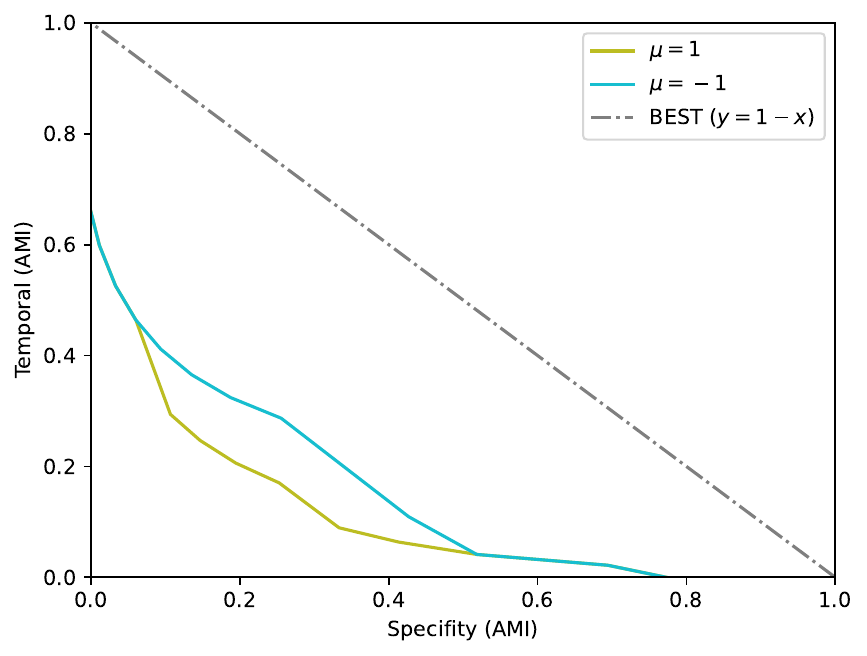}
    \caption{Specify and temporal AMI trade-off when $\mu = \pm 1$ for 1000 samples. $\mu=1$ proposes better trade-off points comparing to $\mu = -1$.}
    \label{fig:change_lambda_AMI_AUC_muVS}
\end{figure}
Furthermore, there is a threshold for the $|\lambda|$, after which the method clusters all nodes into one cluster. That is because if $\lambda$ magnitude is high, the fairness term is dominating term and therefore, it must be zero $z^T s = 0$. A trivial clustering to make $z^T s = 0$ is to put all data points to one cluster (a local optimum; See Figure~\ref{fig:change_lambda_-1_1_observe_AMI_spec_temp}). And due to the discrete and highly non-convex nature of the clustering problem, the algorithm cannot recover from such local optimum. Therefore, it it is crucial to tune $\lambda$ even if a fully balanced (totally fair under demographic parity doctrine) clustering is desired. 

\noindent \textbf{Asymmetric Change of Temporal and Specificity AMI.}
Referring to Figure~\ref{fig:change_lambda_01_03_observe_AMI_spec_temp}, the specificity and temporal AMI changes are not symmetric. The reason is that here, we are continuously solving a discrete problem: assigning labels to graph nodes. Controlled by our method's hyper-parameters, $\mu$, and $\lambda$, each solution for the clustering method corresponds to an assignment of clusters to nodes. The specificity and clustering accuracy trade-off is determined by the shape of the graph and the specificity vector. Therefore, there could be an unreachable trade-off point that no clustering could accomplish.
\end{document}